\documentclass[acmsmall,screen,nonacm]{acmart}

\renewcommand\footnotetextcopyrightpermission[1]{}
\AtBeginDocument{%
  }

\begin{document}

\title{The Disruptive Impact of Large Language Models on Capture the Flag Competitions and the Path Toward Fair Play}

\author{Michael Macaulay}
\authornote{Corresponding author.}
\orcid{0000-0000-0000-0000}
\affiliation{%
  \institution{WMG, University of Warwick}
  \city{Coventry}
  \country{United Kingdom}}
\email{michael.macaulay@warwick.ac.uk}

\author{Harmony Bouabid}
\affiliation{%
  \institution{WMG, University of Warwick}
  \city{Coventry}
  \country{United Kingdom}}
\email{neo.bouabid@warwick.ac.uk}

\author{Ang Guo Gen}
\affiliation{%
  \institution{WMG, University of Warwick}
  \city{Coventry}
  \country{United Kingdom}}
\email{guo-gen.ang@warwick.ac.uk}

\author{Sasha Shaw}
\affiliation{%
  \institution{WMG, University of Warwick}
  \city{Coventry}
  \country{United Kingdom}}
\email{sasha.shaw@warwick.ac.uk}

\renewcommand{\shortauthors}{Macaulay et al.}

\begin{abstract}
Capture the Flag (CTF) competitions are among cybersecurity's most effective training grounds, developing practical skill across cryptography, web exploitation, and binary exploitation. Large language models (LLMs) can now solve a growing share of challenges with minimal human input, raising urgent questions about fairness, the validity of rankings, and whether participation still delivers the learning that justifies the effort. This paper reports a mixed-methods study of LLM impact on modern CTFs, combining a synthesis of published benchmarks, including a recent government evaluation, case studies of live competition across three challenge categories, structured observation of the public channels where the community debates AI use, and semi-structured interviews with experienced players and organisers. We map the current human--machine capability boundary by category, showing that easy and intermediate challenges in cryptography, web, and binary exploitation are now reliably automated while narrower sub-categories continue to resist. We find that community disagreement about whether AI should be permitted is downstream of an undeclared prior question: what a competition is for. Against this backdrop we contribute a four-component safeguard framework, combining tiered competition divisions, LLM-resistant challenge design, telemetry used investigatively, and a draft community code of conduct, together with a decision tool that ties the combination of safeguards to a competition's declared purpose. The argument reaches beyond CTFs to any setting in cybersecurity where a demonstrated result is taken as evidence of an underlying ability.
\end{abstract}



\maketitle

\hypersetup{pdfcreator={},pdfproducer={},pdfpublisher={The Authors}}

\section{Introduction}

We lead the competition programme of a university cyber security society, and much of that work is watching students prepare for Capture the Flag (CTF) competitions. We have watched them learn the hard way: grinding through binary exploitation late into the night, reading disassembly a register at a time, building the kind of intuition that comes only from hours of failing before something clicks. That process is slow and largely invisible from the outside, and in our experience it remains the single most reliable way anyone has ever learned to do this work.

On the other side of the same competitions we now watch what frontier large language models can do to those same challenges, often in seconds: reading a binary, recognising a vulnerability class, writing and iterating on an exploit, and returning a flag before a human has finished reading the prompt. This is not, in itself, a bad thing. The same capability that dissolves a challenge in seconds has also opened the CTF domain to people who would never have got through the door before: newcomers who can now reach a foothold on a problem that would once have ended their attempt entirely, and who learn something real in the process. The technology has widened the on-ramp, and that is a genuine good.

The difficulty is the gap that opens when these two worlds meet on one scoreboard: students who treat a CTF as a genuine test of skill, who did the slow work of becoming good, find themselves up against teams completing the same challenges with AI assistance and none of the hard-earned development behind it. That gap is demoralising to watch. It is also, we have come to think, the more interesting problem, because it is not really about cheating, and it is not unique to CTFs. It is a specific instance of a question now pressing on the whole of our field: how do we assess technical competence when a machine can produce the artefact of competence without the competence itself?

CTFs are an unusually clear place to examine that question, because they have always been an explicit test, a scoreboard, a ranking, a flag either captured or not, and because the community around them has spent the last eighteen months arguing, loudly and publicly, about what that test now measures. The conclusions reach well beyond CTFs, to certification, to recruitment, to university assessment, and to every other context in cybersecurity where a demonstrated result is taken as evidence of an underlying ability. This is not an esoteric concern about a hobby. Industry should care because CTFs are where the next generation of talent hones the skills that firms will be hiring tomorrow; academia should care because CTFs are a large part of what draws talented people to study computing in the first place; and government should care because CTFs are the neutral ground on which the people who will solve tomorrow's national cybersecurity problems are first identified.

Our central argument is that the debate, as it is usually conducted, asks the wrong question. It oscillates between a technical question, is using an LLM a legitimate tool or is it cheating, and an affective one, does it ruin the fun. Both miss the prior question: what is the competition for? A CTF run to identify the fastest path to a flag and a CTF run to develop and certify human skill are two different activities that happen to share a format, and an intervention that is obviously fine in the first is obviously corrosive in the second. The reason the community cannot resolve its argument is that it is trying to answer a single question, is AI allowed, that has no single answer until the purpose is declared. The absence of a declared purpose is not a neutral act of tolerance; in an unregulated competitive environment it quietly resolves itself in favour of whatever maximises the score, which is precisely the outcome that hollows out the test for everyone who came to be measured by it.

We ask three research questions:
\begin{description}
\item[RQ1.] Which CTF challenge categories and sub-categories can current LLMs and agentic systems solve, and where does human-relevant difficulty remain?
\item[RQ2.] How do experienced players and organisers perceive the impact of LLMs on the purpose, fairness, and pedagogical value of CTFs?
\item[RQ3.] What safeguards can organisers combine to preserve the educational value of CTFs, and how should the combination depend on a competition's declared purpose?
\end{description}

We contribute: (a) an empirically grounded map of the current human--machine capability boundary across three CTF categories; (b) qualitative evidence of how the community is negotiating LLM use; and (c) a four-component safeguard framework with a decision tool that ties the choice of safeguards to a competition's declared purpose.

\section{Background and Related Work}

CTFs are structured contests in which participants solve security challenges, spanning cryptography, web exploitation, binary exploitation, reverse engineering and forensics, to retrieve hidden strings called flags. They are widely used in universities and industry to develop and evidence practical skill, and they feed recruitment and national talent pipelines.

\begin{figure}[t]
  \centering
  \includegraphics[width=\linewidth]{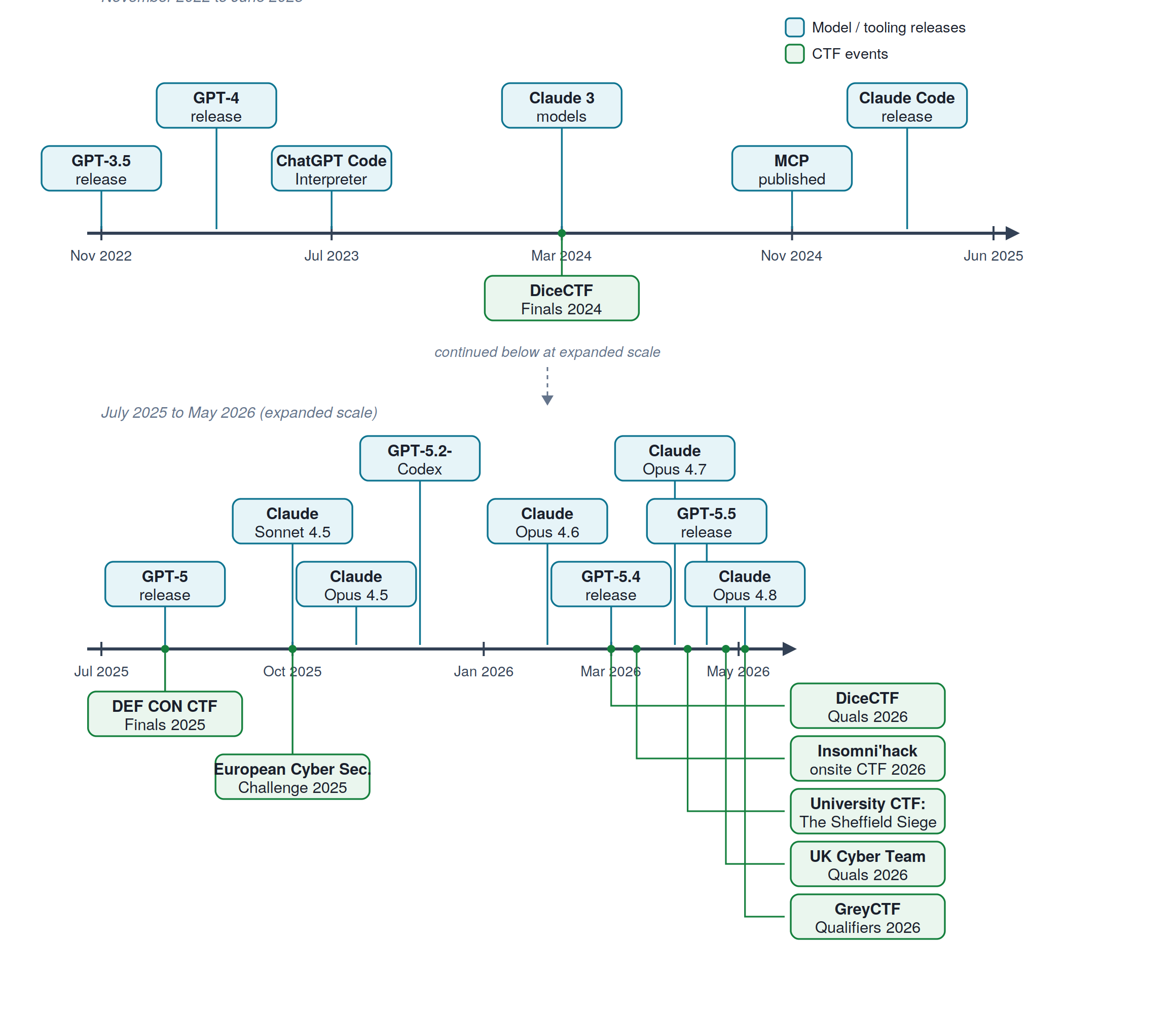}
  \caption{Timeline of the disruption, plotting the model and tooling releases that drove it (above each axis) against the CTF events at which its effects became visible (below), from the public release of ChatGPT in November 2022 to the spring 2026 competitive season. The lower panel repeats the final year at an expanded scale: the clustering of capability-relevant releases from late 2025 onward, and of the events discussed in this paper alongside them, is the chronology the five-phase account in this section describes from the players' side. Event positions in the spring 2026 season are approximate.}
  \Description{A two-panel horizontal timeline. The upper panel spans November 2022 to June 2025 and shows sparse model and tooling releases above the axis (GPT-3.5, GPT-4, ChatGPT Code Interpreter, Claude 3, MCP, Claude Code) and one CTF event below it (DiceCTF Finals 2024). The lower panel repeats July 2025 to May 2026 at an expanded scale, showing a dense cluster of model releases above the axis and five spring-2026 CTF events connected by right-angle leader lines to a stacked column of labels on the right.}
  \label{fig:timeline}
\end{figure}

The disruption is best understood not as a single moment but as a timeline of five overlapping phases (Figure~\ref{fig:timeline}), each marked by what the models could newly do and what previously protected categories began to lose. The first phase opened with the public release of ChatGPT (GPT-3.5) in November 2022, the moment general-purpose models reached players at scale. At first, LLMs were essentially a faster Stack Overflow: useful for boilerplate, regexes, parsers and quick scripts, but not competitors in their own right. In practice this meant pasting a prompt into a web interface and, with luck, getting back a script that ran and produced the flag, but only for rudimentary challenges. This first phase was the period when the community could comfortably treat AI as just another piece of tooling.

Text-heavy categories then began to fall, cryptography first, because the problems were easy to represent directly in language and code; the model was no longer translating a player's thinking into code, it was doing the work. Other categories held out longer for structural reasons worth naming, because they explain why the next phase mattered so much. Binaries were opaque to token-based models: a stripped ELF is not language in any useful sense. Pwn required interactive debugger loops that early models could not drive. And many challenges relied on stripped, packed, or deliberately weird formats that defeated pattern matching. Web had analogous protections: strange application logic, custom stacks, and multi-step exploit chains that were hard for early models to reason through end-to-end. Three distinct protections, then: representational opacity, tool-chain dependency, and challenge-specific quirks.

All three dissolved at once with the agentic transition. Harnesses that let models drive a debugger or browser, inspect output, run tools, read errors and revise payloads meant representational opacity weakened, because the agent could disassemble, decompile and re-pose the problem to itself; tool-chain dependency weakened, because the agent could now use the tool-chain; and challenge-specific quirks weakened, because the loop allowed the model to recover from a wrong first guess rather than committing to it. At one major live event an autonomous solver completed a head-to-head challenge and auto-submitted the flag before its own player realised he had won. The point is not the spectacle but the categorical change in what using AI could mean during a live event.

Two developments through late 2025 and early 2026 turned that categorical change into a routine one. The first was economic: at a major European competition in October 2025, AI was used extensively and openly, and at least one team is reported to have spent on the order of \$8{,}000 in model credits over a single event, an early sign that competing at the top now carries a tooling budget. The second was technical, and is best understood as a change in context windows rather than raw model quality. The first widely used models were limited to roughly 8{,}000 to 16{,}000 tokens, often not enough to hold an entire challenge. By early 2026, coding- and security-specialised models with context windows approaching a million tokens could take in a whole challenge, its source, and a long chain of tool output at once, and the practical reach of the agentic loop expanded sharply as a result. Players widely report a step-change in CTF performance around January and February 2026 that tracks this expansion more closely than it tracks any single new model release.

Organiser awareness has lagged the capability. Ahead of a beginner-oriented university competition, we asked the organisers their AI policy and were told, in good faith, that LLMs were not much help and that no policy was needed. An agent we built and ran afterwards, purely as an experiment, pulled every challenge from the platform's API, spawned a solver per challenge, and solved twenty of them in under ten minutes. We record this not to embarrass anyone but because it is the most honest illustration we have of the central problem: the scene now moves fast enough that, unless you are playing every weekend, it is genuinely easy to underestimate how capable these systems have become. The organiser's no-policy-needed is itself a policy.

Prior work has benchmarked models and agents against public CTF datasets. Cybench~\cite{zhang2025cybench}, NYU CTF Bench~\cite{shao2024nyu} and InterCode~\cite{yang2023intercode} cover hundreds of jeopardy-style tasks and show a consistent pattern: solo-model accuracy is bounded and category-skewed, but agentic harnesses raise the ceiling substantially in every category, and recent agentic systems~\cite{ji2025measuring} reach solve rates that would place an agent in the upper half of a live leaderboard. Conceptually, we draw on the distinction between tools as cognitive prostheses in the extended-mind sense~\cite{clark1998extended} and the offloading of the reasoning itself~\cite{risko2016cognitive}, and on the illusion of competence that unearned success can induce~\cite{kruger1999unskilled}. Pre-LLM tools such as solvers and SMT frameworks were tolerated because the player still identified the vulnerability, picked the tool and chained the exploit; LLM agents rupture that boundary by automating the reasoning, leaving the player a middleman validating output rather than directing it. Our contribution is to connect this capability picture to the purpose of competitions and to organiser-actionable design.

\section{Method}

We used a mixed-methods design with four strands. First, a synthesis of published benchmarks of frontier models and agents against public CTF datasets, supplemented by a recent government evaluation. Second, observational case studies of live competition during the 2025--26 season across three categories, cryptography, web exploitation and binary exploitation, drawing on events including DiceCTF, Insomni'hack and national qualifiers, and on the authors' own competitive play in those categories. Third, structured observation of public community channels (r/securityCTF, r/netsec, the HackTheBox and TryHackMe Discords, and CTFtime threads), used for context only, with no individuals identified. Fourth, three semi-structured interviews (30 to 60 minutes, conducted remotely and audio-recorded with consent) with experienced players and organisers, recruited through a short open call and community networks: a national-team-level player whose home category is cryptography, a player with several years of competitive experience across domestic and international events, and an organiser who designs challenges and advises on competition AI policy. Participants are identified by pseudonym, and identifying contextual details have been generalised or minimised to protect anonymity (see below).

The interview study received full ethical approval from the University of Warwick Biomedical and Scientific Research Ethics Committee (reference BSREC 156/25-26). Written informed consent was obtained from each participant; recordings and the pseudonym key are held separately within the University's approved environment, and identifiable data will be deleted once the analysis is complete.

\paragraph{Protecting participant anonymity}
Because the competitive CTF community is small and closely networked, pseudonymisation alone offers limited protection: a combination of contextual attributes can narrow identity even when a name is withheld. We have therefore generalised or removed identifying details that are not analytically necessary, including participants' nationality, the precise length of their competitive experience, the specific committees or events they are associated with, and any locating particulars of the incidents they describe. Where a distinctive quotation sat alongside a describable event, we have separated the two so that the pair cannot be used to triangulate an individual. No alteration has been made to the substance of what any participant said; only the surrounding identifying context has been reduced. Quotations are reproduced verbatim or lightly paraphrased for readability, and are attributed only by pseudonym.

\paragraph{Limitations}
The interview sample is small and self-selected, and we offer it as illustrative texture for the positions we map rather than as a representative survey; one participant had been given an outline of the study's framing before recording, so that participant's convergence with our central question is best read as informed elaboration rather than as independent arrival at the same view. The case studies are observational, and the capability landscape moves quickly, so the boundary we map is a snapshot whose direction of travel matters more than its exact position. Several of the safeguards we discuss, telemetry above all, rest on methodology that is still immature.

\section{Findings: The Capability Boundary (RQ1)}

Three categories span the disruption curve: cryptography, which fell earliest and has settled; web exploitation, which is mid-transition; and binary exploitation, which held out longest on structural grounds and has given way fastest of all. Table~\ref{tab:boundary} summarises what has fallen and what still holds; the subsections below give the observational detail behind it.

\subsection{Cryptography: the settled case}

An author who plays cryptography competitively reports that LLMs have changed what it means to compete in the category. The 2026 edition of DiceCTF, a prominent international competition, is often cited as the epitome of that change and as proof that the art is dying: teams took first bloods on multiple medium-difficulty challenges and openly credited their LLM agent for chaining the exploit, while the scoreboard showed a steeper-than-usual drop in solve times during the opening minutes. Some called it cheating; others defended it as smart tooling. We argue that this fight goes nowhere because both sides are making a category mistake, swinging between a technical assessment, is it a tool or a cheat, and an affective complaint, does it ruin the fun, neither of which asks the only question that matters: what is the competition actually for?

The reason the debate is polarised is that people are playing two different games on the same scoreboard. The instrumentalist views CTFs as competition-as-outcome: victory is the only goal, prompt engineering is a skill, and inference speed is no different from a fast GPU. The traditionalist views the CTF as competition-as-process, an aret\'e-based arena for cultivating human excellence, on which an LLM solve is a counterfeit achievement because it skips the intellectual struggle required to forge tacit knowledge. Pre-LLM tools such as RsaCtfTool, z3 or angr were tolerated because they acted as cognitive prostheses in the sense Clark and Chalmers~\cite{clark1998extended} describe: they sped up execution, but the player still had to identify the vulnerability, pick the tool, interpret the output and chain the exploit. LLMs rupture that boundary by attempting to automate the reasoning itself, so that the player becomes a middleman validating the machine's output rather than directing it~\cite{risko2016cognitive}. Organisers often stay silent to avoid taking sides, or because any rule is hard to implement online; but silence is not neutral. In an unregulated competitive environment, outcome-maximising strategies always win, and without a declared purpose the pluralism of the game collapses into a single, hollow metric.

Cryptography has taken the hardest hit. Its challenges are frequently open-source, built around human-readable text and code, and easy to test locally, which makes them unusually exposed. Easy and medium challenges revolve around well-known exploits, and these are exactly the patterns an LLM excels at recognising and applying: classical attacks, RSA structural weaknesses, padding-oracle attacks, Wiener's attack, H\aa stad's broadcast, and basic AES misimplementation are now solved end-to-end by frontier agents. What the model adds is the connective tissue that ties a publicly available base exploit to the challenge's specifics and automates the orchestration. At recent qualifier events, LLM-assisted teams first-blooded intermediate classical challenges in times physically implausible for human solvers.

The ceiling, however, is untouched. Complex elliptic-curve isogenies, advanced lattice-based cryptography and bespoke mathematical obfuscations still reliably break LLMs, reflecting what Bender et al.~\cite{bender2021stochastic} call the stochastic-parrots nature of these systems: they recombine statistical patterns from their training data but remain unable to synthesise genuinely novel mathematical logic. When a challenge demands first-principle reasoning, deriving a non-standard algebraic relationship, or exploiting a weird interaction in a custom protocol, they hallucinate or loop, because there is no prior pattern to copy. Manual play retains its purpose here even where the competitive payoff on the easy and middle curve has collapsed: practising attacks by hand, and learning the fine conditions under which each applies, builds the mathematical reflexivity required to invent novel exploits rather than copy old ones, and a category that loses that intuition loses its capacity to advance.

\subsection{Web exploitation: the live case}

Web is the category where this boundary feels most dynamic. Where cryptography has settled into a clear picture of what falls and what holds, web is in the middle of that settling. Classic SQL injection, predictable IDOR, common SSTI patterns, JWT misconfigurations, SSRF-to-metadata chains and basic prototype pollution are increasingly within reach of frontier agents. The important point is not that these models have replaced Burp, sqlmap or existing tooling, but that they provide the connective tissue: recognising the shape of a bug from surface behaviour, selecting the right tool, and chaining several weak signals into a working exploit. That connective work used to be the player's most distinguishing contribution; it is now the part the agent is most reliably good at.

What has held, at least for the moment, is the harder edge of the category: bespoke request-smuggling and desync chains, novel business-logic flaws specific to one application's state machine, race conditions whose difficulty lives in dynamic state rather than in any static pattern, and heavily obfuscated client-side code that resists token-level reasoning. What these have in common is the absence of a written-up precursor the model can recognise; difficulty lives in the live behaviour of the system rather than in a class of bug the model has read about, and as soon as a write-up exists the protection erodes. This is a moving line, not a settled one. At Insomni'hack the top team solved 27 of 30 challenges using AI; the three that resisted are instructive about where the human-only frontier now sits: a roughly eight-gigabyte challenge that had to be distributed physically because the tooling could not ingest it, a video game running on a physical console that a player had to interact with by hand, and one, called Socializer, that required social-engineering the organisers in person. The remaining frontier is moving toward physicality, novelty, human context and messy interaction. Manual exploitation, meanwhile, trains something AI-assisted play does not: an instinct for application surface area, the ability to read an unfamiliar stack, and the patience to chase a hunch that looks stupid for the first three attempts.

\subsection{Binary exploitation: the case that held longest}

An author who came up through pwn and reverse engineering reports where the remaining confidence ended. These were the categories that were supposed to be safe, and for a long time they were, protected by opacity, tool-chain dependency and deliberately strange formats. The 2026 DiceCTF qualifiers is where that confidence ended. DiceCTF is not a soft competition, and its pwn category in particular is regarded as one of the hardest around; in that edition LLMs solved almost the entire pwn set, the only survivor being a difficult Linux kernel exploitation problem and the single hardest challenge in the category. When the floor of too-hard-for-a-model has risen to the top one challenge in a top-tier event, the protection is effectively gone.

What has fallen is most of the category by volume. Frontier models are reliable on the bread-and-butter of pwn, the majority of standard stack and heap exploitation challenges, which should surprise no one: these are the most common challenges in the game, which makes them the most thoroughly written up, and write-ups are exactly what a model trains on. More uncomfortable is browser exploitation, among the hardest things human players do, where models are unexpectedly strong, and the reason is instructive: browsers are a critical, heavily-funded area of live security research, so the same effort that produces the public corpus is, in effect, training the models on the target. The harder humans find a category, the more research attention it attracts, and the more material the model inherits.

The common defensive moves fail, and the reasons matter for anyone designing challenges. Obfuscating binaries, removing source, cranking optimisation and stripping symbols is close to the worst option available: it makes the challenge markedly less fun for human players while having little effect on the solver. Building challenges around large applications fails the same way, because a model scans a codebase far faster than a person reads it, so the size that exhausts a human is an advantage to the machine. And patching real targets like browsers or kernels runs straight back into the corpus problem. A quieter cost matters pedagogically: a model does not only fail loudly, it fails silently, returning a confident wrong answer or a hallucinated assumption the player never sees. A player with real intuition catches it and moves on; a player who has only ever offloaded the reasoning follows the model down a rabbit hole with no instinct that anything is wrong. The faster a player skips the struggle, the less equipped they are for the moment the tool quietly lets them down.

\subsection{Benchmarks corroborate the observational picture}

The case studies above are observational. To ground them empirically we read across the recent literature that has benchmarked frontier models and agents against public CTF datasets. Cybench~\cite{zhang2025cybench}, NYU CTF Bench~\cite{shao2024nyu} and InterCode~\cite{yang2023intercode} together cover hundreds of jeopardy-style tasks and show a consistent pattern: pure language-model accuracy is bounded and category-skewed, with strong performance on text-heavy categories such as cryptography and forensics and weaker performance on pwn and obfuscated reverse engineering, but agentic harnesses raise the ceiling substantially across every category, with the largest gains in web and the smallest in genuinely novel-mathematics cryptography. Recent agentic work, including the CTFAgent system of Ji et al.~\cite{ji2025measuring}, reports solve rates on medium-difficulty jeopardy challenges that would, in a live event, place an agent comfortably in the upper half of the leaderboard.

To the academic benchmarks we add a more recent and, for this argument, more arresting data point. In February 2026 the UK AI Security Institute published an evaluation of frontier models against CTF-style challenges, drawing on public challenge sets aimed at high-school and university students~\cite{aisi2026cyber}. Its most capable models solve close to the whole of the medium-to-hard difficulty band and, at a fixed agentic token budget, a substantial share of the hardest challenges the evaluators could construct, those intended to imitate real-world exploitation. Two things make this significant beyond the headline numbers. First, it is an independent, state-backed measurement rather than a community anecdote, which closes off the objection that the disruption is overstated by the people most invested in it. Second, the Institute reports that the length of cyber task a frontier model can complete at high reliability has been doubling every few months, an unambiguously upward trend that is the property mattering most for any organiser trying to design a competition that will still be meaningful next year rather than only this one.

Two findings from this literature matter for the rest of the paper. First, the gap between solo-model and agentic-harness performance is, in every benchmark, larger than the gap between successive model generations: the agentic transition matters more than model scale, an observation the timeline reaches independently from the players' side. Second, the categories at the top of the difficulty curve in the benchmarks, novel mathematical cryptography, bespoke logic, race conditions, are precisely the sub-categories we identify above as still resisting frontier agents in live play. The observational picture and the benchmark picture agree, which is some reassurance that the boundary we map is real rather than artefactual.

\begin{table}[t]
  \caption{The human--machine capability boundary across three CTF categories, mid-2026. ``Reliably automated'' reflects observed solves by frontier agents in live competition; ``still human-relevant'' identifies sub-categories that continue to resist automation. The boundary is a moving line: resistant designs erode as write-ups enter training corpora.}
  \label{tab:boundary}
  \small
  \begin{tabular}{@{}p{1.7cm}p{1.9cm}p{4.0cm}p{4.0cm}@{}}
    \toprule
    \textbf{Category} & \textbf{Disruption stage} & \textbf{Reliably automated by frontier agents} & \textbf{Still human-relevant}\\
    \midrule
    Cryptography & Settled (fell earliest) & Classical attacks; RSA structural weaknesses (Wiener, H\aa stad); padding-oracle attacks; basic AES misimplementation; orchestration tying known exploits to challenge specifics at solve speeds implausible for humans & Complex elliptic-curve isogenies; advanced lattice-based cryptography; bespoke mathematical obfuscation; challenges requiring first-principles derivation of novel algebraic relationships\\
    \addlinespace
    Web exploitation & Mid-transition (line still moving) & Classic SQL injection; predictable IDOR; common SSTI; JWT misconfiguration; SSRF-to-metadata chains; basic prototype pollution; recognising a bug class from surface behaviour and chaining weak signals into a working exploit & Bespoke request-smuggling and desync chains; novel business-logic flaws specific to one application's state machine; race conditions living in dynamic state; heavily obfuscated client-side code; physical or social components. Protection erodes once a write-up exists\\
    \addlinespace
    Binary exploitation & Recently collapsed (held longest) & The majority of standard stack and heap exploitation; browser exploitation, where heavy public research documentation effectively trains the models on the target & Top-difficulty kernel exploitation (a single challenge survived an otherwise swept top-tier set); niche targets lacking a deep public corpus, though this buys limited time\\
    \bottomrule
  \end{tabular}
\end{table}

The deeper finding is conceptual. The debate is polarised because two games are being played on one scoreboard: an instrumentalist game of competition-as-outcome, on which prompt engineering is a skill, and a traditionalist game of competition-as-process, on which an LLM solve is counterfeit because it skips the struggle that forges tacit knowledge. Without a declared purpose, outcome-maximising strategies win by default, which is where the stakes become pedagogical.

\section{Findings: What the Community Is Negotiating (RQ2)}

\subsection{Five recurring positions}

From observation of the public channels, five positions recur. They are not five answers to one question but answers to subtly different questions, which is part of why the community cannot resolve its argument.

\paragraph{Tool-neutrality} treats AI as simply another tool in an unbroken tradition of external aids: in 2012 an edge meant knowing which debugger plugin to use, in 2018 where to find the right write-up, and today how to prompt effectively; the cognitive work is still human, so restricting AI is no more principled than restricting search engines would have been.

\paragraph{AI-as-cheating} rejects the comparison directly: a search engine surfaces information, but an LLM can construct the solution path, so a flag is meaningful only if the core mental work remains human, and AI use voids that condition regardless of category or context.

\paragraph{Conditional acceptance} permits AI where it mainly explains established material, such as standard web misconfigurations and known cryptographic weaknesses, but not where the task is to reason through genuine novelty, or accepts AI in beginner and educational events while drawing the line at qualifiers and finals where ranking is at stake.

\paragraph{Disclosure-first} holds that the primary problem is not AI use but undisclosed AI use, which distorts the meaning of public write-ups and prevents the community from calibrating what a solve demonstrates, and that mandatory post-competition disclosure would resolve most of the fairness concern without restricting access.

\paragraph{Pay-to-win} is a structural worry that cuts across the others: that modern play is increasingly AI versus AI, and that if the objective is to win, participants are pushed toward the strongest paid model. One player sharpens this into a question of infrastructure rather than subscription: the more tokens a team can spend, and the more it invests in virtual private servers behind the latest models, the faster it solves, to the point that a well-resourced setup can clear a competition in minutes. The economic signal is already visible, with at least one team reported to have spent on the order of \$8{,}000 in model credits at a single recent event. This introduces the possibility of a resource-stratified competitive environment in which access to premium tooling becomes part of competitive performance, and it even raises the prospect, which some organisers may choose to embrace deliberately, of a separate category in which that infrastructure race is the point. Any framework that hopes to be useful must take all four locations, the cognitive core, its context-dependence, honesty about use, and structural access, seriously.

\subsection{What players and organisers say}

The three interviews populate that map rather than resolving it. Elmer, a national-team-level cryptography player, occupies the AI-as-cheating position but holds it with a precision the bare label hides: he draws the line not at AI but at agents. Agents, he said, ruin the competition, while AI used to search for information or explain a concept, the background work one would previously have done through a search engine, only goes faster. This is almost word for word the boundary the paper draws between a cognitive prosthesis and cognitive outsourcing, and it is evidence that the distinction is one some players already make for themselves. He also describes the lived cost of the shift: where it was once possible to jump into a CTF and feel like you had a fighting chance, he now finds that everything already has dozens of solves, which he calls very demotivating, a collapse of morale rather than a drift in difficulty. Asked what a CTF is for, he reached unprompted for the paper's organising question, arguing that a competition should first and foremost be about learning, and that when it becomes more about the competition than the learning, the plot has been lost. His organiser-side experience carries the same message: helping to run an event for younger entrants where the organisers deliberately fielded a constrained model, able to explain concepts but not to identify flaws or solve challenges, he observed that many teams attempted the AI route, failed, and disengaged rather than turning to the challenge itself, a pattern of learned helplessness that bears directly on the pedagogical cost.

Jammy, a player with several years of competitive experience, gives the instrumentalist drift a first-person shape, and from someone candid that it has happened to him. He still competes, but mostly to qualify rather than to learn, and he uses AI extensively when he does, autosolving, completing exploit scripts and re-prompting for hints, while still working by hand in the categories he knows best. He is direct about the effect on his own development: having meant to take up cryptography, he found the model could do it all and so he did not really need to learn it. If a competition rewards only the flag and not the working, he argues, players will gravitate to the most efficient path to it, with costs not only to technical skill but to the softer disciplines of teamwork and of communicating an idea clearly. He points to a human-only division at one event that drew little interest because nothing rewarded entering it, and reads the scene as contracting, with AI accelerating rather than causing the decline.

Hani, a challenge designer who advises on competition AI policy, supplies the organiser's seat and a less pessimistic reading. He reports the same widening gap between AI-reliant juniors and pre-LLM seniors, alongside a blunt governance fact: senior players have told organisers they will not compete unless AI is banned. He describes a graduated compromise he considers workable if imperfect, in which AI use is prohibited during the main body of a competition but permitted in a limited closing phase under monitored, reduced-points conditions. Yet he is the most accommodating of the three, framing AI as just a tool, likening a ban to banning calculators, and observing that the youngest entrants treat the model as a normal collaborator, to the point that one of his most popular challenges asked players to prompt-engineer an AI rather than avoid it. His worry is not that AI is used but that juniors skip the unglamorous hours that build deep understanding, a restriction he frames as protective of learning rather than of rankings.

These interviews do not resolve the disagreement the factional map sets out; they populate it. What is striking is less where the three disagree than where they converge: asked what should be done, all three return, by different routes, to what the competition is for. Elmer states it as a principle, Jammy as a diagnosis, that if an event rewards only the answer and not the working, the answer is all players will pursue, and Hani as a governance question organisers must now actually decide. That convergence is the empirical counterpart of this paper's central claim: the community's workable disagreements are downstream of a prior question about purpose that most competitions have not yet declared.

\section{The Pedagogical Stakes}

The case studies set up an uncomfortable question. If the easy and middle curve of two of the largest CTF categories can now be solved end-to-end by frontier agents, what is the point of a beginner still working through them by hand? This is the gap named in the introduction, examined from inside the categories where it bites hardest. It is also where the authors of this paper most visibly disagree about emphasis, and we have chosen to preserve that disagreement rather than smooth it, because the two halves of the gap are both real.

\subsection{The cost to hard-won skill}

Most discussion of LLMs in CTFs has compared two experienced ideals, the player who uses AI and the player who does not. The more important comparison is between newcomers who learn the field with AI assistance from the start and those who learned it without. When an LLM solves a classical RSA challenge before the beginner understands why that weakness exists, the flag arrives without the struggle: a short-term reward that supplies almost no skill or knowledge, tending to induce the illusion of competence that Kruger and Dunning~\cite{kruger1999unskilled} describe and masking an atrophy of first-principle reasoning. The deeper concern is structural rather than moral. Cryptography as a field arrived where it is because of human intuition; a category that loses the players who would have developed that intuition loses, over time, the capacity to advance, and the same logic applies to pwn and to anything else in cyber whose hard problems require the slow accumulation of low-level instincts.

An author who has played since 2014 adds a dimension easy to miss if the argument stays inside the competition. The skills a CTF develops are precisely the ones rarely taught formally and that transfer directly to defending real systems. A challenge is often a real vulnerability in disguise: a bug-bounty hunter or vulnerability researcher who found something in production recreates it inside a container so that others can learn the trick by exploiting it themselves. The player who works it out by hand, on the debugger, asking on IRC or Discord when stuck, waiting for a write-up or messaging the author when they fail, internalises not just one bug but a class of bug, and security researchers spend their careers finding that same cluster recurring across different products. CTFs, on this view, are not a game that happens to resemble the work; they are where the field quietly trains the people who will later secure the software the rest of us depend on. The instrumental drift toward prompt-and-flag does not just produce weaker competitors; it threatens the mechanism by which curiosity-driven problem-decomposers, the benign kind of hacker, are made at all.

\subsection{The gain in accessibility}

The pessimistic case has weight, but it is not the whole picture, and a player who entered CTF competitions during the LLM era holds primary evidence about the on-ramp that no veteran can give credibly. At Break The Syntax CTF 2026, a participant working through a pwn challenge used an AI assistant to reason through a shellcode constraint problem they had not encountered before. The challenge created a read-write-execute memory mapping and called it as shellcode, but restricted the bytes that could be copied to printable characters, a constraint that collapses to a known alphanumeric-shellcode encoding problem only if the player already knows that technique class exists. For an occasional player without that prior exposure, the constraint would most likely have ended the attempt entirely. The interaction did not produce the flag; it named the constraint class, pointed toward the relevant encoding approach, and gave the player a conceptual foothold they reported not having had before. That is the distinction this argument turns on: not AI as a solution engine, but AI as the mechanism that kept a newer player in the problem long enough for learning to occur.

Three early jeopardy challenges illustrate the same dynamic across categories. In a binary task where a tampered OpenSSH binary concealed a password encoded with a single-character XOR cipher keyed on the password length, the assistant identified the encoding class, explained why XORing the ciphertext with its own length recovers the plaintext, and named the idiom to execute it, understanding that would not have emerged from staring at the binary diff alone. In a web task involving a PHP file loader vulnerable to directory traversal, it identified the pattern and explained why the server failed to sanitise the file-path input, so the category of vulnerability became meaningful rather than accidental. And in a forensics task where a valid PNG contained a 7z archive appended after the IEND marker, it explained how the PNG specification defines file boundaries and why parsers stop reading at IEND regardless of trailing bytes, which made the hexdump behaviour immediately legible. Across all three the AI did not replace the work; it named the technique class, explained the underlying logic, and gave enough structure that the solution path became followable.

The on-ramp AI provides is real, but it has a shape, and that shape has edges. During a maritime OSINT challenge requiring multi-source correlation across vessel-tracking records, dock coordinates and Haversine distance calculations, an assistant asked to reason through the correlation produced a structured, confident and wrong reconstruction of the vessel's route, because it was pattern-matching against general maritime knowledge rather than reasoning from the specific data supplied. The error was not immediately obvious; catching it required returning to the raw data, cross-referencing the coordinates by hand, and running the calculation manually to confirm the proposed route was geometrically inconsistent with the source. That process of verification, which would not have happened had the answer been accepted at face value, was the most instructive part of the challenge, and it established a working norm: AI output in data-grounded tasks is a hypothesis to be checked, not a result to be submitted. The judgement to recognise that distinction is not something the model can supply; it must be built through exactly this kind of correction. AI widens the on-ramp, but it does not replace the effort of walking up it.

\subsection{Holding both halves together}

The two sub-sections above are the same phenomenon viewed from the experienced and the newer seat. The accessibility gain is real, and it is the most significant widening of the CTF on-ramp the field has ever seen. The pedagogical cost is also real, and it falls hardest on exactly those players for whom the on-ramp is widest, because they are the ones most likely to mistake conceptual exposure for procedural mastery. The two are not in opposition; they are two halves of a single gap, and any credible position on AI in CTFs has to hold both at once. The strongest case for optimism does not rest on the claim that AI always improves participation; it rests on the narrower claim that AI can widen access while preserving meaningful learning, provided players remain responsible for interpretation, verification and judgement. Whether players stay on the on-ramp or step off it is a question of practice norms, not tooling, which is why the paper's final answer is a framework rather than a verdict.

\section{Organiser Responses and a Safeguard Framework (RQ3)}

\subsection{What competitions have tried}

\begin{figure}[t]
  \centering
  \includegraphics[width=0.86\linewidth]{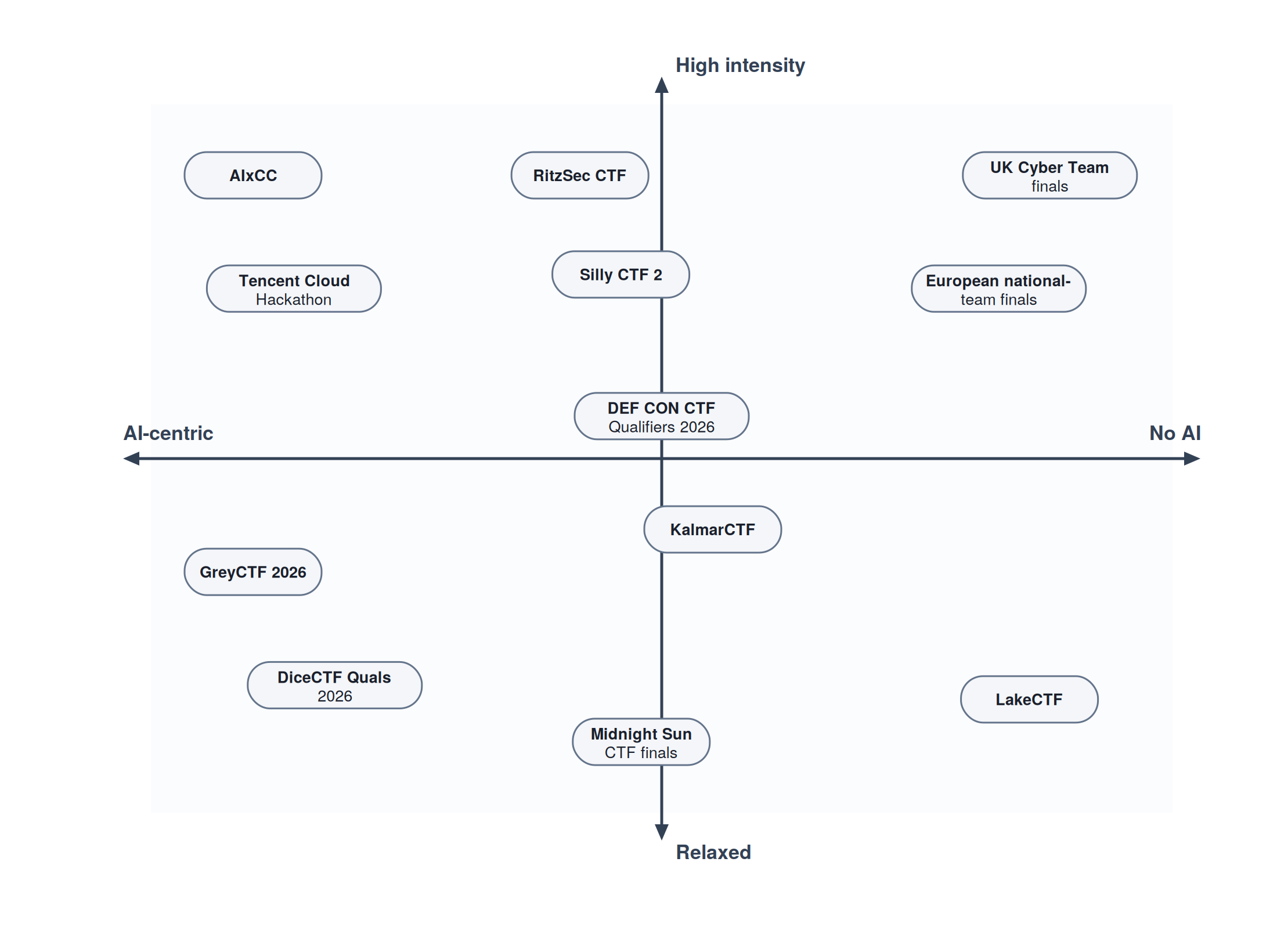}
  \caption{An indicative map of existing approaches, locating events by their AI policy on the horizontal axis (from AI-centric on the left to no-AI on the right) and by competition intensity on the vertical axis. In the AI-centric, high-intensity corner sit the autonomous-AI contests and AI-first hackathons; across the top centre, events investing in LLM-resistant design; in the no-AI, high-intensity corner, proctored national-team finals; near the centre of the policy axis, large qualifiers running declared-AI taxonomies; and the lower, more relaxed band runs from permissive events with no enforced policy through community finals to strict in-person events. Positions are approximate and reflect the events' stated or observed postures during the 2025--26 season. The map is a description of where events currently sit, not a recommendation; the framework in Section~\ref{sec:framework} and the decision tool in Section~\ref{sec:decisiontool} are what convert a position on it into a deliberate choice.}
  \Description{A two-axis scatter map. The horizontal axis runs from AI-centric on the left to No AI on the right; the vertical axis runs from high intensity at the top to relaxed at the bottom. Twelve CTF events are placed as labelled pills across the four quadrants, ranging from AI-first hackathons in the top left to proctored national-team finals in the top right, with qualifiers near the centre and more relaxed community events along the bottom.}
  \label{fig:policymap}
\end{figure}

Surveying events that have published or described an AI stance, two axes capture most of the variation (Figure~\ref{fig:policymap}). The first is permissiveness: how much AI use the rules allow, from competitions that actively encourage it to those that ban it outright. The second is enforcement: how hard the organisers actually work to make the policy bite, from those who step back entirely to those who proctor, lock down and inspect. Almost every real competition sits in the interior of that square rather than at a corner, and plotting them there makes two things visible: that a stated policy and an enforced policy are different objects, and that the most common failure is not a permissive policy but a strict policy with no enforcement behind it. Within that space, four response types recur, with a fifth that inverts the question entirely.

\paragraph{Network restriction and air-gapped finals}
The strictest response removes the tool by removing its access. At several national finals the posture is essentially that of an exam: no AI, with proctors walking the room and ejection for anyone caught. A European national side and the UK Cyber Team have both run finals on this model, and the UK variant pushed it furthest, distributing locked-down virtualised laptops on which players could not even copy and paste. Even then it was not airtight: one competitor managed to reach an AI feature through a search interface and was caught by network inspection, itself an illustration that air-gapping in practice is a layered effort rather than a single switch. LakeCTF applied the same severity to an in-person event with a lighter touch culturally, banning not only AI but search engines and even editor tab-completion, and its organisers came away optimistic. The common lesson is encouraging for finals and unhelpful everywhere else: air-gapping works when everyone is in one room and does nothing for the online qualifiers where ranking is actually established and where AI use is therefore most consequential.

\paragraph{AI-permitted-but-declared, and tiered side-boards}
A second cluster permits AI but tries to make its use legible, either by declaration or by carving out a protected division. KalmarCTF ran a separate self-declared scoreboard for players competing without AI; the design was sound in principle but carried no prizes for the human board, precisely the missing incentive one interviewee described. Midnight Sun's own finals drew a finer line, allowing AI to be consulted for general understanding and advice but not fed challenge text or used to solve a challenge directly, a distinction close to the prosthesis-versus-agent boundary, though to the best of our knowledge not actively proctored. DEF CON CTF Qualifiers 2026 took the permissive-but-bounded route at scale, permitting AI but forbidding its use to solve a challenge outright, and its declared-AI taxonomy of no-AI, low-AI and human-led gave players an explicit channel to self-classify. At the permissive, hands-off end sit events such as Grey CTF, which openly encouraged AI use without enforcement, and DiceCTF 2026, where the absence of a policy functioned as one: competing seriously effectively required AI. The recurring weakness of the whole cluster is the boundary problem: declared tracks help culturally, by giving the honest somewhere to stand, but the moment a competitor at the line treats declaration as optional, the only thing that can tell is behavioural detection.

\paragraph{LLM-resistant challenge design}
A third response tries to build challenges humans can solve but models cannot, targeting the category-specific weaknesses set out above. Two flavours are worth distinguishing. The first is genuine LLM-resistant design: RitzSec and Midnight Sun both invested real effort here, and RitzSec's organisers reported writing a small number of challenges that AI could not solve and that withstood the attempt in practice. The second is the trap. Silly CTF experimented with a range of novel approaches, the most discussed of which was the LLM honeypot: a challenge whose human-readable text instructs the reader not to submit, warning that doing so will result in a ban, before supplying a flag anyway. A human reads the warning and stops; an agent tasked only with extracting and submitting flags walks straight into it. It is an elegant idea and an effective one, but, as its designers acknowledge, something of a one-trick approach that loses its bite once the pattern is known. Both flavours share the structural limitation set out in the case studies: LLM-resistant design works for a cycle or two, until the idea is written up and absorbed into the next round of training data. Today's LLM-resistant design is tomorrow's training data.

\paragraph{Telemetry and behavioural detection}
The fourth response targets behaviour rather than tooling, and the most instructive worked example to date is RitzSec, an online event whose organisers permitted limited AI, forbade fully AI-solved challenges, and then actually enforced the rule, banning on the order of a hundred teams after the competition closed. They did so not from any single signal but from a combination: prize-eligible teams were required to submit write-ups, which cannot be produced convincingly by a team that did not understand its own solves; the top teams were additionally interviewed about the hardest challenges; teams that submitted hallucinated flags, the kind a model invents and a human almost never produces, were flagged immediately; and solve timings were examined, but deliberately used only to raise suspicion and direct attention rather than as evidence for a ban in itself. The RitzSec case is the strongest existing demonstration of telemetry done responsibly, precisely because no individual signal was treated as dispositive.

\paragraph{Removing the human: the autonomous-AI frontier}
A fifth route does not try to manage AI use at all but inverts the contest around it. Events in the lineage of large autonomous-security competitions, such as the AI-versus-AI challenge run at DEF CON, ask entrants to submit agentic frameworks that must find and patch vulnerabilities with no human in the loop, with prize pools that dwarf conventional CTFs. This is not a safeguard against the disruption this paper describes; it is a different competition that has declared its purpose unambiguously to be the machine's capability rather than the human's. We include it because it is the cleanest illustration of the paper's thesis from the opposite direction: an event that openly says it is about AI faces none of the fairness crisis that afflicts events which leave the question undeclared. Pay-to-win is not a distortion of such an event; it is the event.

Running underneath the four management responses is a calculation each player makes, and bringing it into view explains why enforcement posture cannot be uniform across an event. From the seat of a player tempted to break the policy, the decision weighs reward against risk: the reward is the competitive advantage AI confers and whatever it unlocks, which at the top of the field can be considerable, and the risk is the probability of being caught multiplied by the consequences. In person, detection probability is high, because someone can look over a shoulder or ask a pointed question, which is why air-gapped finals hold up. Online, detection probability collapses, which is why the same nominal policy that works in a final is close to unenforceable in a qualifier. The uncomfortable corollary is that in online qualifiers the incentive structure quietly favours the rule-breaker unless the organiser has deliberately raised the probability of detection or lowered the reward. A real positioning of any competition therefore has a third dimension behind the two axes: not just how permissive and how enforced, but how strong the temptation is in the first place.

None of these responses is sufficient on its own. The pattern across the events that have held up best is not adoption of any single response but combination of several, weighted according to the kind of competition being run and matched to where the temptation actually bites. This is the evidential basis for the framework that follows: a single intervention cannot work, but a coordinated set of partial interventions, declared in advance and matched to purpose, can.

\subsection{Four components, combined by purpose}
\label{sec:framework}

We propose not a solution but a framework: four components an organiser can combine, in proportions determined by the purpose they have declared. The framework is deliberately neither a recommendation to ban nor a recommendation to permit. It is a structure for making the choice coherent, so that a competition's rules about AI follow from what it is actually trying to measure rather than from whatever the organiser could most easily enforce.

\paragraph{Tiered competition divisions}
The most direct way to stop two incompatible games being scored on one leaderboard is to give them separate leaderboards. A formally designated AI-permitted division lets newer players use the on-ramp without stigma, and without distorting the ranking in an open or invitational division where AI use is restricted. The value of tiering is not that it catches anyone but that it removes the incentive to misrepresent, because there is a transparent division for every kind of participant. The Kalmar side-board in Section~\ref{sec:framework} carries the warning that comes with it: a protected division only works if there is a reason to enter it, and a human-only board with no prizes attached, as both that side-board and the event one interviewee described demonstrate, will draw few takers however principled its design. Tiering refuses the premise that the field must choose between access and integrity, but only if the incentives are arranged so that each division is worth competing in.

\paragraph{LLM-resistant challenge design}
The case studies are, read from the designer's side, a specification. The crypto and web sub-categories that still reliably defeat frontier agents, novel mathematical structure, bespoke protocol composition, race conditions, genuinely new logic flaws, and challenges whose difficulty lives in dynamic state rather than static pattern, are a map of where human-relevant difficulty currently sits. Designing toward those properties keeps a category meaningful for human players, at least for a cycle. The RitzSec and Midnight Sun slop-resistant challenges and Silly CTF's honeypot are proof that the design space is real and that determined organisers can win a round in it. The caveat is that this protection is temporary; challenge design is therefore best understood not as a fix but as a recurring cost, a garden that has to be re-planted every season, and organisers should budget for it as such rather than expecting a design that holds indefinitely.

\paragraph{Telemetry-based detection heuristics}
Solve telemetry, timing patterns, the shape of a solve path, the signature of an automated loop, is in principle the most promising long-term mechanism, because it targets behaviour rather than the tool and so does not depend on the tool being declared. But a population trained in offensive security is exactly the population most able to defeat behavioural monitoring: prompts can be retyped by hand, pages saved and worked offline, timing artefacts smoothed deliberately. The methodological maturity to distinguish AI-assisted from skilled-but-fast human play is not yet there, and an organiser who treats telemetry as a primary enforcement mechanism risks spending real resources on a signal they cannot trust. Our position is that telemetry belongs in the framework as a supporting, investigative instrument, useful for flagging anomalies for human review, not for adjudicating them, exactly as the RitzSec organisers used timing data, and that any organiser adopting it should be candid, with themselves and with competitors, about how weak the signal currently is.

\paragraph{A draft community code of conduct}
The first three components are things organisers do. The fourth is something the community adopts, and it has the longest reach, because it addresses what we identify as the real fairness concern: not AI use, but undisclosed AI use. A norm of post-competition disclosure, stating which tools were used and, where meaningful, which prompts proved instructive, turns AI from a hidden advantage into a shared educational resource and lets the community calibrate what a given solve demonstrates. A code of conduct cannot be enforced the way a network restriction can; it works, when it works, by becoming the default that serious participants expect of one another, and it takes hold only when anchored by institutions the community already trusts. There is a second lever organisers should not overlook: sponsors. Many people now in industry leadership came up through this community, and a sponsor can make support conditional on a declared policy or on the existence of a genuinely resourced human division, and in doing so shift norms faster than any single organiser acting alone.

\subsection{A decision tool: what is your competition for?}
\label{sec:decisiontool}

\begin{figure}[t]
  \centering
  \includegraphics[width=\linewidth]{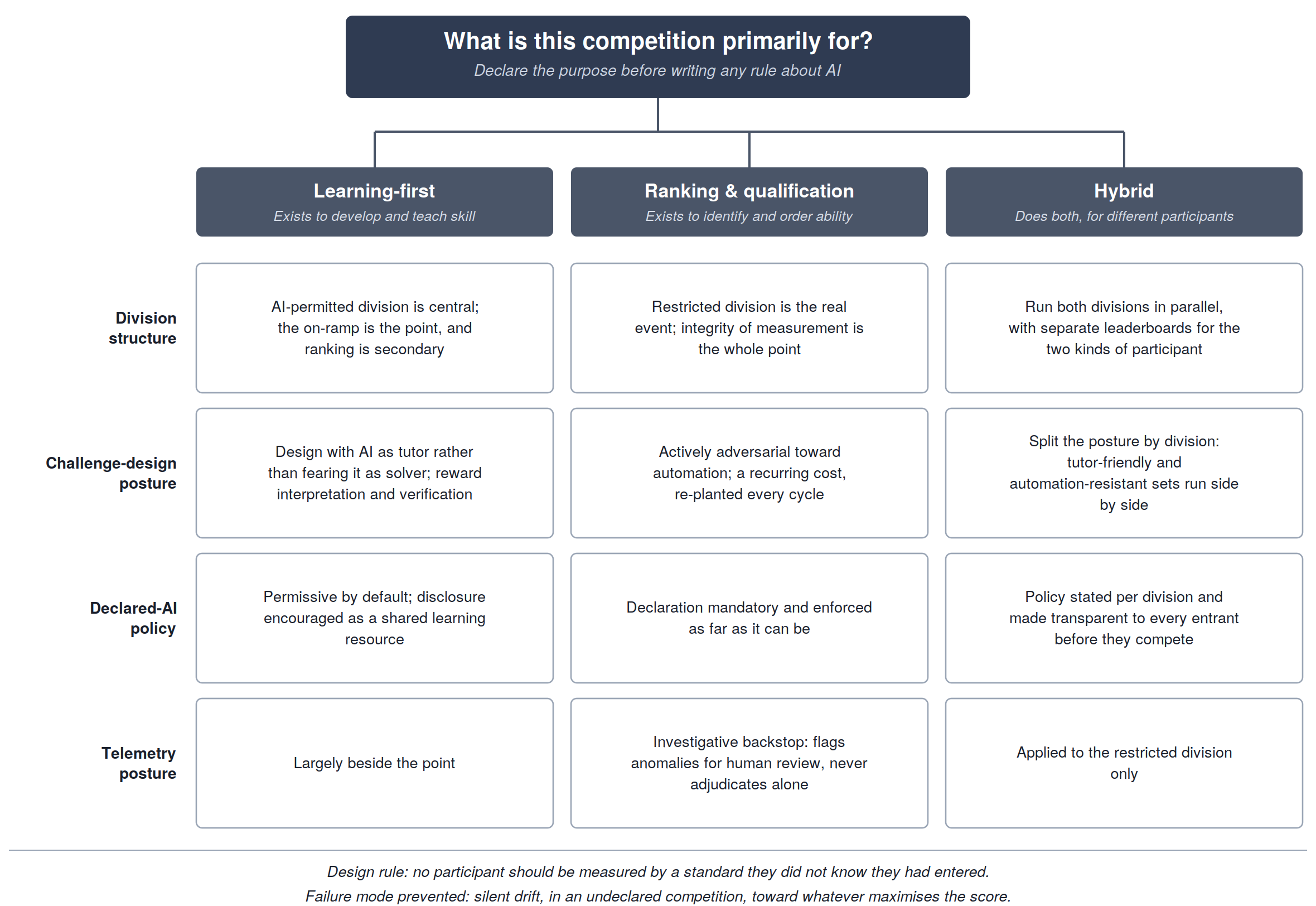}
  \caption{Decision tool for competition organisers. The root question, what is this competition primarily for, branches into three competition types, learning-first, ranking-and-qualification, and hybrid, each resolving the four safeguard-framework components (division structure, challenge-design posture, declared-AI policy, and telemetry posture) with different weightings. The tool takes no position on whether AI use is good or bad; it prevents a single failure mode, the silent drift, in an undeclared competition, toward whatever maximises the score.}
  \Description{A decision-tree diagram. A dark box at the top asks ``What is this competition primarily for?'' and branches into three column headers: learning-first, ranking and qualification, and hybrid. Below each column, four rows (division structure, challenge-design posture, declared-AI policy, telemetry posture) give the recommended posture for that competition type. A footer states the design rule that no participant should be measured by a standard they did not know they had entered.}
  \label{fig:decisiontree}
\end{figure}

A framework with four components still leaves the organiser with a choice about how to combine them, and that choice cannot be made well without first answering the question this paper opened with. We propose answering it with a simple decision tree (Figure~\ref{fig:decisiontree}), not because the matter is simple but because forcing the question to the front, before any rule about AI is written, is what makes every subsequent choice coherent. The tree begins with the root question and branches into three kinds of competition, because in practice almost every event is one of three things. A learning-first event exists to develop and teach skill; its participants are there to get better, and ranking is secondary. A ranking-and-qualification event exists to identify and order ability, often as a gate to something else, and here the integrity of the measurement is the whole point. And a hybrid event, which most real competitions are, tries to do both at once for different participants. The organiser's first task is simply to say, openly, which of the three they are running. A fourth kind deserves naming for completeness, the autonomous-AI contest that declares the agent itself the competitor; it sits outside the tree precisely because it has already answered the root question, and the fairness crisis the tree exists to prevent never arises for it.

From that root, each branch resolves into the same four sub-decisions, drawn from the framework but weighted differently according to purpose. A learning-first event should lean hard on the AI-permitted division and on challenge design that uses AI as a tutor rather than fearing it as a solver; declared-AI policy here is permissive by default, and telemetry is largely beside the point. A ranking-and-qualification event inverts those weights: here the restricted division is the real event, challenge design is actively adversarial toward automation, declaration is mandatory and enforced as far as it can be, and telemetry, with all the caveats above, earns its place as an investigative backstop. A hybrid event has to do the hardest thing of all, running both postures simultaneously and being transparent with participants about which posture applies to which division, so that no one is measured by a standard they did not know they had entered. The risk-and-reward analysis sharpens the same advice: the more a branch depends on enforcement, the more it should reckon honestly with how much weaker that enforcement is online than in person, and design the reward structure accordingly.

It is worth being equally clear about what the decision tree does not do. It does not resolve the conceptual debate about what AI use means; it is consistent with several different views. It does not prescribe the community norms; those are the community's to negotiate, not the organiser's to impose. And it takes no position on whether AI use is good or bad. It asks only that the organiser take a position, declare a purpose, and make their choices follow from it. The single failure mode it is designed to prevent is the one we identified at the start: the silent drift, in an undeclared competition, toward whatever maximises the score, and the quiet hollowing-out of the test that follows.

\section{Discussion and Conclusion}

Our findings carry implications beyond CTFs. Wherever cybersecurity and computing education take a demonstrated result as evidence of underlying ability, coursework, practical examinations, certification, portfolio hiring, the same capability boundary and the same purpose question apply, and the framework's logic, declare the purpose and then choose safeguards that follow from it, transfers directly. The limitations noted in the method apply: the interview sample is illustrative rather than representative, the boundary is a snapshot whose direction of travel matters more than its exact position, and several safeguards, telemetry above all, rest on immature methodology. Evaluating the framework in a live competition is the natural next step, and one we intend to pursue through the competition programme we run.

We frame this as a path toward fair play rather than a defence of the status quo. Fairness here does not mean banning the technology, and it does not mean pretending the old way of competing can be preserved unchanged. It means competitions being honest about what they measure, so that the people who enter them know what they are being measured on, and so that the hard-won skill we still believe is worth having retains a place where it can be developed, demonstrated and recognised for what it is. The question is not finally about CTFs at all. It is about whether, across our field, we can keep telling the difference between competence and its convincing imitation. Capture the Flag is simply where that question has arrived first, and most clearly, and where, if we are willing to declare what our tests are for, we still have time to answer it well.

\begin{acks}
We thank the interview participants for their time and candour, and the student members of the Warwick Cyber competition programme whose practice and questions shaped this work.
\end{acks}

\section*{Ethics and Privacy Statement}

The interview component of this study received full ethical approval from the University of Warwick Biomedical and Scientific Research Ethics Committee (reference BSREC 156/25-26). Participants gave written informed consent and are identified only by pseudonym; because the competitive community is small, identifying contextual details have been generalised to reduce re-identification risk, without altering the substance of what participants said. The paper documents capabilities that could, in principle, inform attempts to game competitions; we judge the benefit of an open, shared account of the current boundary, and of a framework that helps organisers protect the educational value of their events, to outweigh that risk, particularly as the underlying capabilities are already widely known within the community. No personal data beyond the pseudonymised interview material was collected.

\section*{Use of Generative AI}

Generative AI is both a subject and, in a bounded way, an instrument of this study, and we distinguish the two. As study material, documented interactions with AI assistants form part of the reported evidence in Section~6: the illustrative walkthroughs of specific challenges were conducted deliberately and are described as data, and the authors take full responsibility for their interpretation. As a writing aid, the authors used an AI assistant to support drafting, structuring, and copy-editing of the manuscript; all such output was reviewed, verified, and edited by the authors, who take full responsibility for the content, including all claims, citations, and the accuracy of every reference. AI tools were not used to process participants' personal data at any stage.

\section*{Funding}

This research received no external funding.

\section*{Data Availability}

The interview data underlying Section~5 are pseudonymised and access-restricted under the terms of the study's ethical approval and participant consent, and cannot be shared openly; the pseudonym key and audio recordings are held securely and will be deleted once the analysis is complete. The benchmark and competition evidence synthesised in Sections~2 and~4 is drawn from publicly available sources cited in the references. No new quantitative dataset was generated.

\section*{Author Contributions}

The authors declare the following contributions using the CRediT taxonomy. \emph{Conceptualization:} all authors. \emph{Methodology:} all authors. \emph{Investigation:} all authors, comprising the category case studies (cryptography, web exploitation, and binary exploitation), the community observation, and the semi-structured interviews. \emph{Formal analysis:} all authors. \emph{Writing, original draft:} all authors. \emph{Writing, review and editing:} all authors. \emph{Project administration and supervision:} the corresponding author, who leads the competition programme from which the study arose.

\bibliographystyle{ACM-Reference-Format}
\bibliography{ctf-refs}

\end{document}